# Maximum Uncertainty Procedures for Interval-Valued Probability Distributions

Michael Pittarelli
Computer Science Department
State University of New York College of Technology
Utica, NY 13504-3050

## Abstract

Measures of uncertainty and divergence are introduced for interval-valued probability distributions and are shown to have desirable mathematical properties. A maximum uncertainty inference procedure for marginal interval distributions is presented. A technique for reconstruction of interval distributions from projections is developed based on this inference procedure.

## 1. Interval-Valued Distributions

An interval-valued distribution is a finite sequence of intervals $([Pl_1, Pu_1], \ldots, [Pl_n, Pu_n])$ such that $0 \leq Pl_j \leq Pu_j \leq 1$, $\sum_{j=1}^{n} Pu_j \geq 1$, and $\sum_{j=1}^{n} Pl_j \leq 1$. Interval distributions generalize real-valued probability distributions and arise naturally in many situations. They may represent collections of confidence intervals derived from frequency data, imprecisely stated subjective probabilities, known linear equality or inequality constraints, etc. Thus, interval distributions sometimes provide a more realistic characterization of uncertainty than do real-valued probability distributions. In addition, certain practical problems, e.g., inconsistency of overlapping probability estimates over structured spaces, are much less severe for such distributions [Pittarelli, 1989] At the same time, interval distributions can be put to many of the same uses as real-valued distributions. For example, criteria have been developed for decision analysis using probability intervals [Seidenfeld, 1983; Loui et al., 1986; Pittarelli, 1989].

To illustrate a situation in which available information may be summarized by an interval-valued distribution, consider the probabilistic database [Cavallo and Pittarelli, 1987; Pittarelli, 1989] (collection of marginal distributions) $D = \{p_1, p_2\}$:

| X | $p_1(.)$ | Y | $p_2(.)$ |
|---|---|---|---|
| $x_1$ | 0.7 | $y_1$ | 0.6 |
| $x_2$ | 0.3 | $y_2$ | 0.4 |

The *extension*, $E(D)$, of this database is the (infinite) set of all solutions $p: \{x_1, x_2\} \times \{y_1, y_2\} \to [0,1]$ to the system of equations:

$p(x_1 y_1) + p(x_1 y_2) = 0.7$

$p(x_2 y_1) + p(x_2 y_2) = 0.3$

$p(x_1 y_1) + p(x_2 y_1) = 0.6$

$p(x_1 y_2) + p(x_2 y_2) = 0.4$

$p(.) \geq 0.$

For any database D, lower and upper bounds on joint probabilities $p(s_j)$, as p ranges over $E(D)$, may be calculated via linear programming as $\min_{p \in E(D)} p(s_j)$ and $\max_{p \in E(D)} p(s_j)$. Since the function $f_j: E(D) \to [0,1]$, where $f_j(p) = p(s_j)$, is a linear function of a convex set, its range is also convex and is thus a subinterval of $[0,1]$:
$$f_j(E(D)) = [\min_{p \in E(D)} p(s_j), \max_{p \in E(D)} p(s_j)].$$
The unknown joint probability $p(s_j)$ can take on any value in this range. This is all that one can say about $p(s_j)$ without taking into account interactions between its value and those of the other $p(s_i)$.

For the database above, the joint probability intervals
$$i_D(s_j) = [\min_{p \in E(D)} p(s_j), \max_{p \in E(D)} p(s_j)]$$
are

| X | Y | $i_D(.)$ |
|---|---|---|
| $x_1$ | $y_1$ | [0.3, 0.6] |
| $x_1$ | $y_2$ | [0.1, 0.4] |
| $x_2$ | $y_1$ | [0, 0.3] |
| $x_2$ | $y_2$ | [0, 0.3] |

Any $p \in E(D)$ is included in $i_D$, in the sense that $p(s) \in i_D(s)$ for all $s \in \{x_1, x_2\} \times \{y_1, y_2\}$. $i_D$ is the most informative (i.e., narrowest) interval distribution for which this is guaranteed to be the case.

*Definition:* For two interval distributions i and i' over a set of n-tuples T, i is *more informative than* i', denoted $i \leq i'$, iff $i(s) \subseteq i'(s)$ for all $s \in T$.

Identifying a real-valued distribution p over T with a degenerate interval distribution, i.e., $p(s) = [p(s), p(s)]$, any $p \in E(D)$ is more informative than the interval distribution $i_D$. For example, $E(D)^*$, the maximum entropy element of $E(D)$, is more informative than $i_D$:

| X | Y | $E(D)^*(.)$ |
|---|---|---|
| $x_1$ | $y_1$ | 0.42 |
| $x_1$ | $y_2$ | 0.28 |
| $x_2$ | $y_1$ | 0.18 |
| $x_2$ | $y_2$ | 0.12 |

Unless additional constraints are known (e.g., probabilistic independence of attributes) or it happens that $E(D) = \{E(D)^*\}$, such precision is unwarranted.

Suppose now that information is available in the form of an interval-valued probabilistic database; i.e., a collection $I = \{i_1, \ldots, i_m\}$ of interval distributions such that $i_j: S_j \to P[0,1]$, where $S_j = \times_{v \in V_j} \text{dom}(v) = \text{dom}(V_j)$. The task is to infer interval probabilities for events $s \in \text{dom}(V_1 \cup \cdots \cup V_m)$.

*Example:* $I = \{i_1, i_2\}$ is

| X | Y | $i_1(.)$ | Y | Z | $i_2(.)$ |
|---|---|---|---|---|---|
| $x_1$ | $y_1$ | [0.2, 0.6] | $y_1$ | $z_1$ | [0, 0.3] |
| $x_1$ | $y_2$ | [0.4, 0.8] | $y_1$ | $z_2$ | [0.2, 0.5] |
| $x_2$ | $y_1$ | [0, 0.2] | $y_2$ | $z_1$ | [0.1, 0.4] |
| $x_2$ | $y_2$ | [0, 0.1] | $y_2$ | $z_2$ | [0, 0.2] |

$V_1 = \{X,Y\}$, $V_2 = \{Y,Z\}$, $dom(X) = \{x_1, x_2\}$, etc. It is desired to find interval probabilities for $s \in dom(\{X,Y,Z\}) = \{x_1 y_1 z_1, x_1 y_1 z_2, ..., x_2 y_2 z_2\}$.

An interval database may be regarded as providing constraints on real-valued probabilities of joint events.

*Definition:* The *real-valued extension*, $R(I)$, of an interval database $I = \{i_1, \ldots, i_m\}$ is the set of real-valued distributions over $dom(V_1 \cup \cdots \cup V_m)$ satisfying the inequality constraints implied by I.

For $I = \{i_1, i_2\}$ above, $R(I)$ is the set of solutions p to

$$p(x_1 y_1 z_1) + p(x_1 y_1 z_2) \geq 0.2$$
$$p(x_1 y_1 z_1) + p(x_1 y_1 z_2) \geq 0.6$$

$$\vdots$$

$$p(x_1 y_2 z_2) + p(x_2 y_2 z_2) \geq 0$$
$$p(x_1 y_2 z_2) + p(x_2 y_2 z_2) \leq 0.2$$
$$p(x_1 y_1 z_1) + p(x_1 y_1 z_2) + \cdots + p(x_2 y_2 z_2) = 1$$

*Definition:* Let $E(I)^*$ denote the (unique) interval distribution whose components have as endpoints the minimum and maximum values for corresponding probabilities of the elements of $R(I)$. (These are obtainable via linear programming.) Then the *(interval-valued) extension* of I is the set $E(I) = \{i | p \leq i \leq E(I)^*, \text{ for some } p \in R(I)\}$. For the example, $E(I)^*$ is the distribution

| X | Y | Z | $E(I)^*$ |
|---|---|---|---|
| $x_1$ | $y_1$ | $z_1$ | [0,0.3] |
| $x_1$ | $y_1$ | $z_2$ | [0,0.5] |
| $x_1$ | $y_2$ | $z_1$ | [0.2,0.4] |
| $x_1$ | $y_2$ | $z_2$ | [0,0.2] |
| $x_2$ | $y_1$ | $z_1$ | [0,0.2] |
| $x_2$ | $y_1$ | $z_2$ | [0,0.2] |
| $x_2$ | $y_2$ | $z_1$ | [0,0.1] |
| $x_2$ | $y_2$ | $z_2$ | [0,0.1] |

Just as $E(D)^*$ is the maximum uncertainty element of $E(D)$, with uncertainty measured as Shannon entropy, $E(I)^*$ is the maximum uncertainty element of $E(I)$ under various measures of uncertainty appropriate for interval distributions.

## 2. Measures of Uncertainty and Divergence

Let $I^n$ denote the set of all n-component interval distributions. A measure of uncertainty for distributions $i \in I^n$ is a function $u: I^n \to \mathbb{R}$. It seems reasonable to require of any such measure u that

$$i \leq i' \text{ implies } u(i) \leq u(i'),$$

where (the left-hand) $\leq$ is the "more informative than" relation defined above. A simple measure with this property is the function
$$u_o(i) = 1/n \sum_{j=1}^{n} (\max_{x \in i(s_j)} x - \min_{x \in i(s_j)} x).$$
Distance measures currently in favor for reconstruction of real-valued probability and possibility distributions from their projections [Klir and Folger, 1988] have the property that
$$d(x,x^*) = u(x^*) - u(x),$$
where x is the original (joint) distribution and $x^*$, the reconstructed (joint) distribution, is the maximum uncertainty element, according to a suitable uncertainty measure u, of the set of joint distributions compatible with the projections. This may be used as a guide in constructing a distance measure between interval distributions.

Projection of an interval distribution reduces to projection (marginalization) of real-valued distributions as a special case (degenerate intervals). Endpoints of the components of the resulting interval projection are the minimum and maximum values of the components of the real-valued projections of real-valued distributions compatible with the given interval distribution. For the distribution $E(I)^*$ above, its projection onto the set of variables $\{X,Z\}$, denoted $\pi_{\{X,Z\}}(E(I)^*)$, is determined from the system of inequalities with solution set L

$$p(x_1 y_1 z_1) + p(x_1 y_2 z_2) - p(x_1 z_1) = 0$$
$$\ldots$$
$$p(x_2 y_1 z_2) + p(x_2 y_2 z_2) - p(x_2 z_2) = 0$$
$$p(x_1 y_1 z_1) \geq 0$$
$$p(x_1 y_1 z_1) \leq 0.3$$
$$\ldots$$
$$p(x_2 y_2 z_2) \geq 0$$
$$p(x_2 y_2 z_2) \leq 0.1$$
$$p(x_1 z_1) + \cdots + p(x_2 z_2) = 1$$

as $\pi_{\{X,Z\}}(E(I)^*)(x_1 z_1) = [\min_{p \in L} p(x_1 z_1), \max_{p \in L} p(x_1 z_1)] = [0, 0.5]$, etc.:

| X | Z | $\pi_{\{X,Z\}}(E(I)^*)$ |
|---|---|---|
| $x_1$ | $z_1$ | [0,0.5] |
| $x_1$ | $z_2$ | [0,0.7] |
| $x_2$ | $z_1$ | [0,0.3] |
| $x_2$ | $z_2$ | [0,0.3] |

(Notice that the endpoints of the marginals are not obtained simply by adding the endpoints of the corresponding joint intervals.)

When an interval-valued joint distribution i defined on variables V is projected onto a database scheme $X = \{V_1, \ldots, V_m\}$ to form a database $I = \pi_X(i) = \{\pi_{V_1}(i), \ldots, \pi_{V_m}(i)\}$, the (unique) maximum uncertainty element (with respect to $u_0$) of the extension $E(I)$ is $E(I)^*$. Let
$$u_o(E(\pi_X(i))^*) - u_o(i)$$
measure the information loss when i is replaced by its projections onto X. A distance

282

measure between arbitrary pairs of interval distributions in $I^n$ that reduces to this measure of information loss as a special case is

$$d_o(i,i') = 1/n \sum_{j=1}^{n} (|\max_{x \in i(s_j)} x - \max_{x \in i'(s_j)} x| + |\min_{x \in i(s_j)} x - \min_{x \in i'(s_j)} x|).$$

For any i and its reconstruction $i^* = E(\pi_X(i))^*$,

$$d_o(i,i^*) = 1/n \sum_{j=1}^{n} (|\max_{x \in i(s_j)} x - \max_{x \in i^*(s_j)} x| + |\min_{x \in i(s_j)} x - \min_{x \in i^*(s_j)} x|)$$

$$= 1/n \sum_{j=1}^{n} ((\max_{x \in i^*(s_j)} x - \max_{x \in i(s_j)} x) + (\min_{x \in i(s_j)} x - \min_{x \in i^*(s_j)} x))$$

$$= 1/n \sum_{j=1}^{n} ((\max_{x \in i^*(s_j)} x - \min_{x \in i^*(s_j)} x) - (\max_{x \in i(s_j)} x - \min_{x \in i(s_j)} x))$$

$$= 1/n \sum_{j=1}^{n} (\max_{x \in i^*(s_j)} x - \min_{x \in i^*(s_j)} x) - 1/n \sum_{j=1}^{n} (\max_{x \in i(s_j)} x - \min_{x \in i(s_j)} x)$$

$$= u_o(i^*) - u_o(i).$$

It is easy to see that $d_o$ is *nondegenerate*

$$d_o(i,i') = 0 \text{ iff } i = i'$$

and *symmetric*

$$d_o(i,i') = d_o(i',i).$$

Since $|a-c| \leq |a-b| + |b-c|$ for all $a,b,c \in [0,1]$, $d_o$ satisfies the *triangle inequality* also. For all $i_1, i_2, i_3 \in I^n$,

$$d_o(i_1, i_3) \leq d_o(i_1, i_2) + d_o(i_2, i_3).$$

Thus, $d_o$ is a metric distance for $I^n$ and the pair $(I^n, d_o)$ is a metric space of interval-valued distributions.

A set-theoretic relation among database schemes that is important for various types of structural analysis of data [Cavallo and Klir, 1981; Edwards and Havranek, 1985] is the refinement relation.

*Definition*: A database scheme X is a *refinement* of scheme Y, denoted $X \leq Y$, iff for each $V_x \in X$ there exists a $V_y \in Y$ such that $V_x \subseteq V_y$. For example, $\{\{A\},\{B,C\}\}$ is a refinement of $\{\{A,B\},\{B,C\},\{A,C\}\}$.

From:
(1) $X \leq Y$ implies $E(\pi_Y(i))^* \leq E(\pi_X(i))^*$;
(2) $d_o(i, E(\pi_Z(i))^*) = u_o(E(\pi_Z(i))^*) - u_o(i)$; and
(3) $i \leq i'$ implies $u_o(i) \leq u_o(i')$;

it follows that $d_o$ is monotonic with respect to the database scheme refinement relation:

$$X \leq Y \text{ implies } d_o(i, E(\pi_Y(i))^*) \leq d_o(i, E(\pi_X(i))^*).$$

That is, the distance measure $d_o$ has the additional intuitively pleasing property that the information loss measured by means of it never decreases when a database scheme is replaced by a more refined scheme.

*Example*: Consider the interval distribution i defined as

| A | B | C | i(.) |
|---|---|---|---|
| 0 | 0 | 0 | [0.24, 0.26] |
| 0 | 0 | 1 | [0.24, 0.26] |
| 0 | 1 | 0 | [0.04, 0.06] |
| 0 | 1 | 1 | [0.04, 0.06] |
| 1 | 0 | 0 | [0.04, 0.06] |
| 1 | 0 | 1 | [0.04, 0.06] |
| 1 | 1 | 0 | [0.14, 0.16] |
| 1 | 1 | 1 | [0.14, 0.16] |

Scheme {{A,B},{B,C}} is superior to scheme {{A,C},{B,C}} with respect to information loss when i is projected onto either scheme and reconstructed. The relevant projections of i are

| A | B | $i_{\{A,B\}}(.)$ | B | C | $i_{\{B,C\}}(.)$ | A | C | $i_{\{A,C\}}(.)$ |
|---|---|---|---|---|---|---|---|---|
| 0 | 0 | [0.48, 0.52] | 0 | 0 | [0.28, 0.32] | 0 | 0 | [0.28, 0.32] |
| 0 | 1 | [0.08, 0.12] | 0 | 1 | [0.28, 0.32] | 0 | 1 | [0.28, 0.32] |
| 1 | 0 | [0.08, 0.12] | 1 | 0 | [0.18, 0.22] | 1 | 0 | [0.18, 0.22] |
| 1 | 1 | [0.28, 0.32] | 1 | 1 | [0.18, 0.22] | 1 | 1 | [0.18, 0.22] |

The reconstructions of i from schemes {{A,B},{B,C}} and {{A,C},{B,C}} are

| A | B | C | $E(\pi_{\{\{A,B\},\{B,C\}\}}(i))^*$ | A | B | C | $E(\pi_{\{\{A,C\},\{B,C\}\}}(i))^*$ |
|---|---|---|---|---|---|---|---|
| 0 | 0 | 0 | [0.16, 0.32] | 0 | 0 | 0 | [0.06, 0.32] |
| 0 | 0 | 1 | [0.16, 0.32] | 0 | 0 | 1 | [0.06, 0.32] |
| 0 | 1 | 0 | [0, 0.12] | 0 | 1 | 0 | [0, 0.22] |
| 0 | 1 | 1 | [0, 0.12] | 0 | 1 | 1 | [0, 0.22] |
| 1 | 0 | 0 | [0, 0.12] | 1 | 0 | 0 | [0, 0.22] |
| 1 | 0 | 1 | [0, 0.12] | 1 | 0 | 1 | [0, 0.22] |
| 1 | 1 | 0 | [0.06, 0.22] | 1 | 1 | 0 | [0, 0.22] |
| 1 | 1 | 1 | [0.06, 0.22] | 1 | 1 | 1 | [0, 0.22] |

$d_o(i, E(\pi_{\{\{A,B\},\{B,C\}\}}(i))^*) = 0.12$ and $d_o(i, E(\pi_{\{\{A,C\},\{B,C\}\}}(i))^*) = 0.21$. So, using the criterion of minimum information loss when i is reconstructed as $E(I)^*$, measured by $d_o$ (or, equivalently, the maximum $u_o$-uncertainty criterion), {{A,B},{B,C}} would be judged the better scheme.

At this point in the development of a theory of interval distributions, it is difficult to interpret the statement

$$d_o(i, E(\pi_X(i))^*)^*) < d_o(i, E(\pi_Y(i))^*)$$

other than literally; i.e., when i is projected onto X and reconstructed as $E(\pi_X(i))^*$, less information is lost, as measured by $d_o$, than when i is projected onto Y and reconstructed as $E(\pi_Y(i))^*$. This is not completely without value (e.g., for rapid transmission or compact storage of an approximate characterization of i). On the other hand, when such relations hold for schemes of real-valued distributions, they may be interpreted as indicating the relative strength of probabilistic dependencies among the variables of V.

In a previous paper [Cavallo and Pittarelli, 1987], the *approximate probabilistic multivalued dependency* $U \rightarrow\rightarrow_p W$, where $U, W \subseteq V$, is quantified as

$$H(W|U) - H(W|U \cup Z),$$

where $H(x|y)$ is conditional Shannon entropy and $Z = V - U - W$. The smaller this difference,

the stronger is the dependency $U \to\to_p W$; i.e., the less additional information about variables in W can be determined from $U \cup Z$ than from U alone. It is shown that for a scheme X of the form $\{U \cup W, U \cup Z\}$,
$$d(p, E(\pi_X(p))^*) = H(W|U) - H(W|U \cup Z),$$
where d is relative entropy (directed divergence, cross-entropy). Thus, a scheme $X = \{G \cup H, G \cup (V-G-H)\}$ is superior to a scheme $Y = \{I \cup J, I \cup (V-I-J)\}$ of a real-valued p over V, with respect to information loss, iff $G \to\to_p H$ is stronger than $I \to\to_p J$.

When $U \to\to_p W$ reaches full strength, and $H(W|U) - H(W|U \cup Z) = 0$, then p is reconstructable from $X = \{U \cup W, U \cup Z\}$. The full-strength multivalued dependency $U \to\to_p W$ (and perfect reconstructability) coincides with *conditional probabilistic independence* [Dawid, 1979] of W and Z, given U. Random variables W and Z are conditionally independent, given U, iff
$$p(WZ|U) = p(W|U) \times p(Z|U).$$
For any p defined over $V = Z \cup (W \cup U)$,
$$p(WZU) = p(WZ|U) \times p(U).$$
The distribution p is (perfectly) reconstructable from $\{U \cup W, U \cup Z\}$ iff
$$p(WZU) = p(W|U) \times p(UZ) \quad [\text{Lewis}, 1959]$$
$$= p(W|U) \times p(Z|U) \times p(U).$$
Then $p(WZ|U) = p(W|U) \times p(Z|U)$; i.e., W and Z are conditionally independent, given U.

The interval distribution i in the example above was derived from a distribution p as $i(s) = [p(s) - 0.01, p(s) + 0.01]$, where p is recoverable (exactly) via maximum entropy reconstruction from projections onto $\{\{A,B\},\{B,C\}\}$ but not $\{\{A,C\},\{B,C\}\}$. Thus, it is not surprising that the projection-extension information loss criterion for interval distributions identifies $\{\{A,B\},\{B,C\}\}$ as a better model of (scheme for) i than $\{\{A,C\},\{B,C\}\}$. But a satisfactory interpretation in terms of interval-valued dependency concepts is lacking. There has been some work done in the general area of developing interpretations for interval probability statements [Grosof, 1986], but more is necessary.

The measure of uncertainty $u_0: I^n \to \mathbb{R}$ captures one of two distinct aspects of uncertainty associated with an interval distribution, namely, uncertainty regarding a real-valued distribution compatible with it. It does not measure uncertainty regarding the events in the space dom(V) over which a distribution in $I^n$ is defined. The measure
$$u_1(i) = \max_{p \leq i} H(p),$$
where H is Shannon entropy, characterizes this type of uncertainty. This measure also satisfies the condition $i \leq i'$ implies $u(i) \leq u(i')$. However, the maximum $u_1$-uncertainty element of an extension E(I) is not necessarily unique. For example, if E(I) contains the uniform distribution $(1/n, 1/n, ..., 1/n)$ then the set $S \subseteq E(I)$, where $S = \{i \in E(I) | (1/n, 1/n, ..., 1/n) \leq i\}$, contains the maximum $u_1$-uncertainty elements of E(I). But maximizing $u_0$-uncertainty within E(I) also amounts to maximizing $u_1$-uncertainty, since $i \leq E(I)^*$ for all $i \in E(I)$ and $i \leq i'$ implies $u_1(i) \leq u_1(i')$. Any set of interval distributions that possesses a unique maximum element $i^*$ under the "less informative than" relation (as does E(I)) is such that $i^*$ maximizes both $u_0$ (uniquely) and $u_1$ (not necessarily uniquely). Alternatively, perhaps it would be reasonable to look at the *minimum* entropy of any real-valued distribution compatible with an interval

distribution i:

$$u_2(i) = \min_{p \leq i} H(p).$$

Maximizing $u_2$ would be a type of maximin procedure for entropy. Since for $u_2$ as well $i \leq i'$ implies $u(i) \leq u(i')$, $E(I)^*$ also maximizes $u_2$-uncertainty within $E(I)$.

### 3. Summary and Conclusion

Algebraic operations for interval-valued probability distributions are defined on the basis of which a procedure for reconstructing such distributions from their projections is developed. A measure of divergence for interval distributions is necessary to characterize the degree to which a distribution is reconstructable from its projection onto a particular database scheme. A measure that is a metric distance and has an additional important monotonicity property relative to the structure of the lattice of all database schemes over a given set of variables is proposed.

Two types of uncertainty measures for interval distributions are discussed: those that measure uncertainty regarding real-valued distributions constrained by the intervals, and those that measure more directly uncertainty regarding the events in the space over which the interval distribution is defined.